\documentclass{article}

     \usepackage[final]{neurips_2019}

\usepackage[utf8]{inputenc} 
\usepackage[T1]{fontenc}    
\usepackage{url}            
\usepackage{booktabs}       
\usepackage{amsfonts}       
\usepackage{nicefrac}       
\usepackage{microtype}      
\usepackage{verbatim}

\usepackage{commands}

\title{Communication-efficient Distributed SGD with Sketching}

\newcommand*\samethanks[1][\value{footnote}]{\footnotemark[#1]}
\author{%
Nikita Ivkin \thanks{ equal contribution} \ 
\thanks{This work was done while the author was at Johns Hopkins University.}
\\
Amazon \\
 \texttt{ivkin@amazon.com}
\And
Daniel Rothchild \samethanks[1]\\
UC Berkeley\\
\texttt{drothchild@berkeley.edu}
\And
Enayat Ullah \samethanks[1] \\
Johns Hopkins University \\
\texttt{enayat@jhu.edu}
\And
Vladimir Braverman  \thanks{This work was done, in part, while the author was visiting the Simons Institute for the Theory of Computing.}\\
Johns Hopkins University \\
\texttt{vova@cs.jhu.edu}
\And
Ion Stoica \\
UC Berkeley \\
\texttt{istoica@berkeley.edu}
\And
Raman Arora \\
Johns Hopkins University \\
\texttt{arora@cs.jhu.edu}
}

\begin{document}

\maketitle

\begin{abstract}
Large-scale distributed training of neural networks is often limited by network bandwidth, wherein the communication time overwhelms the local computation time. Motivated by the success of sketching methods in sub-linear/streaming algorithms, we introduce \ssgd \footnote{Code is available at \href{https://github.com/dhroth/sketchedsgd}{\texttt{https://github.com/dhroth/sketchedsgd}}}, an algorithm for carrying out distributed SGD by communicating sketches instead of full gradients. We show that \ssgd  has favorable convergence rates on several classes of functions. When considering all communication -- both of gradients and of updated model weights -- \ssgd reduces the amount of communication required compared to other gradient compression methods from $\mathcal{O}(d)$ or $\mathcal{O}(W)$ to $\mathcal{O}(\log d)$, where $d$ is the number of model parameters and $W$ is the number of workers participating in training. We run experiments on a transformer model, an LSTM, and a residual network, demonstrating up to a 40x reduction in total communication cost with no loss in final model performance. We also show experimentally that \ssgd scales to at least 256 workers without increasing communication cost or degrading model performance.
\end{abstract}

\section{Introduction}

Modern machine learning training workloads are commonly distributed across many machines using data-parallel synchronous stochastic gradient descent.
At each iteration, $W$ worker nodes split a mini-batch of size $B$; each worker computes the gradient of the loss on its portion of the data, and then a parameter server sums each worker's gradient to yield the full mini-batch gradient. After using this gradient to update the model parameters, the parameter server must send back the updated weights to each worker. We emphasize that our method can naturally be extended to other topologies as well (e.g. ring, complete, etc.) -- in particular we would then communicate sketches over a minimum spanning tree of the communication graph. However, for ease of exposition, in this work we focus exclusively on the star topology.
For a fixed batch size $B$, the amount of data each worker processes -- and therefore the amount of computation required -- is inversely proportional to $W$. On the other hand, the amount of communication required per worker is independent of $W$. 
Even with optimal interleaving of the communication and computation, the total training time is at least the maximum of the per-worker communication time and per-worker computation time. Increasing the number of workers $W$ therefore yields an increasingly marginal reduction in the training time, despite increasing the overall training cost (number of machines times training time) linearly in $W$.

Several approaches address this issue by using a large batch size to increase the per-worker computation time \citep{you2017large, goyal2017accurate}.
However, theoretical and empirical evidence both suggest that there is a maximum mini-batch size beyond which the number of iterations required to converge stops decreasing, and generalization error begins to increase \citep{ma2017power,li2014efficient, golmant2018computational, shallue2018measuring, keskar2016large,hoffer2017train}.
In this paper, we aim instead to decrease the communication cost per worker.
We use a technique from streaming algorithms called sketching, which allows us to recover favorable convergence guarantees of vanilla SGD. In short, our algorithm has workers send gradient sketches of size $\mathcal{O}(\log d)$ instead of the gradients themselves. Although other methods for reducing the communication cost exist, to our knowledge ours is the only one that gives a per-worker communication cost that is sub-linear in $d$ and constant in $W$. In practice, we show that our method achieves high compression for large $d$ with no loss in model accuracy, and that it scales as expected to large $W$. 

\section{Related Work}

Most existing methods for reducing communication cost in synchronous data-parallel distributed SGD either quantize or sparsify gradients. A number of quantization methods have been proposed. These methods either achieve only a constant reduction in the communication cost per iteration \citep{wen2017terngrad,bernstein2018signsgd1}, or achieve an asymptotic reduction in communication cost per iteration at the expense of an equal (or greater) asymptotic increase in the number of iterations required \citep{alistarh2017qsgd}. Even in the latter case, the total communication required for all of training sees no asymptotic improvement.

Other methods sparsify the gradients instead of quantizing each gradient element \citep{stich2018sparsified,alistarh2018convergence,lin2017deep}.
A popular heuristic is to send the top-$k$ coordinates of the local worker gradients and then average them to obtain an approximate mini-batch gradient. These methods can achieve good performance in practice, but they suffer from a few drawbacks. They currently have no convergence guarantees, since the estimated mini-batch gradient can be very far from the true mini-batch gradient (unless explicitly assumed, as in e.g. \citet{alistarh2018convergence}), which precludes appealing to any known convergence result.
Another drawback is that, although these methods achieve high compression rates when the workers transmit gradients to the parameter server, the return communication of the updated model parameters grows as $\mathcal{O}(W)$: the local top-$k$ of each worker may be disjoint, so there can be as many as $kW$ parameters updated each iteration. This $\mathcal{O}(W)$ communication cost is not just a technicality, since reducing the back-communication to $\mathcal{O}(k)$ would require sparsifying the sum of the local top-$k$, which could hinder convergence. Because of this scaling, local top-$k$ methods suffer from poor compression in settings with large $W$.

From another standpoint, all gradient compression techniques yield either biased or unbiased gradient estimates. A number of quantization methods are crafted specifically to yield unbiased estimates, such that the theoretical guarantees of SGD continue to apply \citep{alistarh2017qsgd,wen2017terngrad}.
However, even without these guarantees, a number of methods using biased gradient estimates were also found to work well in practice \citep{bernstein2018signsgd1,seide20141,strom2015scalable}. Recently, \cite{stich2018sparsified,karimireddy2019error} gave convergence guarantees for this kind of biased compression algorithm, showing that accumulating compression error locally in the workers can overcome the bias in the weight updates as long as the compression algorithm obeys certain properties.
Our method falls into this category, and we prove that compressing gradients with sketches obeys these properties and therefore enjoys the convergence guarantees in \cite{stich2018sparsified}. In effect, we introduce a method that extends the theoretical results of \cite{stich2018sparsified} from a single machine to the distributed setting.
Concurrently with this work, \cite{koloskova2019decentralized} also introduce a distributed learning algorithm with favorable convergence guarantees, in which workers communicate compressed gradients over an arbitrary network topology.

Prior work has proposed applying sketching to address the communication bottleneck in distributed and Federated Learning \citep{konevcny2016federated,jiang2018sketchml}. However, these methods either do not have provable guarantees, or they apply sketches only to portions of the data, failing to alleviate the $\Omega(Wd)$ communication overhead.
In particular, \citet{konevcny2016federated} propose ``sketched updates" in Federated Learning for structured problems, and 
\citet{jiang2018sketchml} introduce a range of hashing and quantization techniques to improve the constant in $\bigO{Wd}$. 

Another line of work that we draw from applies sketching techniques to learning tasks where the model itself cannot fit in memory \citep{aghazadeh18mission, tai2018sketching}. In our setting, we can afford to keep a dense version of the model in memory, and we only make use of the memory-saving properties of sketches to reduce communication between nodes participating in distributed learning.

\section{Preliminaries}
\label{sec:prelim}
\paragraph{SGD.}
Let $\w \in \R^d$ be the parameters of the model to be trained and $f_i(\w)$ be the loss incurred by $\w$ at the $i^{\text{th}}$ data point $(\x_i,y_i)\sim \cD$.
The objective is to minimize the generalization error  $f(\w) = \Eu{(\x_i,y_i)\sim \cD}{f_i(\w)}$. In large-scale machine learning, this objective is typically minimized using mini-batch stochastic gradient descent: given a step size $\eta_t$, at each iteration, $\w$ is updated as $\w_{t+1} = \w_{t} - \eta_t \g_{t},$
where $\g_t=\nabla_\w \sum_{i\in \cM} f_i(\w)$ is the gradient of the loss computed on a minibatch $\cM$. If $\cM$ is randomly selected, then the gradient estimates $\g_t$ are unbiased: i.e. $\E{g_t|\{\w_i\}_{i=0}^{t-1}} = \nabla f(\w_{t-1})$.
As is standard, we further assume that the $\g_t$ have bounded moment and variance: $\E{\norm{\g_t}^2_2 |\{\w_i\}_{i=0}^{t-1} } \leq G^2$ and $ \E{\norm{\g_t - \nabla f(\w_t)}^2_2 |\{\w_i\}_{i=0}^{t-1} } \leq \sigma^2$ for constants $G$ and $\sigma$. 
We adopt the usual definitions for smooth and strongly convex functions:
\begin{definition}[Smooth strongly convex function]
$f:\R^d \rightarrow \R$ is a $L$-smooth and $\mu$-strongly convex if the following hold $\ \forall \ \w_1, \w_2 \in \R^d$,
\begin{CompactEnumerate}
    \item $ \norm{\nabla f(\w_2) - \nabla f(\w_2)} \leq L \norm{\w_2 - \w_1}$ (Smoothness)
    \item $f(\w_2) \geq f(\w_1) + \ip{\nabla f(\w_1)}{\w_2 - \w_1} + \frac{\mu}{2}\norm{\w_2 - \w_1}^2$ (Strong convexity)
\end{CompactEnumerate}

\end{definition}
For smooth strongly convex functions, SGD converges at a rate of $\bigO{\frac{G^2L}{\mu T}}$ \citep{rakhlin2012making}.
\vspace{-10pt}
\paragraph{Count Sketch.} 
Our primary interest is in finding large coordinates (or ``heavy hitters'') of a gradient vector $\g \in \mathbb{R}^d$. Heavy hitter sketches originated in the streaming model, where the vector $\g$ is defined by a sequence of updates $\{(i_j, w_j)\}_{j=1}^{n}$, such that the $j$-th update modifies the $i_j$-th coordinate of $\g$ as $\g_{i_j} \text{ += } w_j$ \citep{charikar2002finding, cormode2005improved, braverman2017bptree}. In the streaming model, sketches must use memory sublinear in both~$d$ and~$n$.

In this work we compress a gradient vector $\g$ into a sketch $S(\g)$  of size $O(\frac{1}{\varepsilon} \log d)$ using a Count Sketch~\citep{charikar2002finding}. A Count Sketch $S(\g)$ approximates every coordinate of $\g$ with an $\ell_2$ guarantee: it is always possible to recover $\hat \g_i$ from $S(\g)$ such that $\g_i^2 - \varepsilon \|\g\|^2_2 \leq \hat\g_i^2\leq \g_i^2 + \varepsilon\|\g\|^2_2$. In addition, $S(\g)$ can approximate the $\ell_2$ norm of the entire gradient. These two properties let a sketch find every $\ell_2$ heavy hitter, i.e. every coordinate $i$ such that $\g_i^2 > \varepsilon \|\g\|^2_2$. With a small enough $\varepsilon$, the set of heavy hitters can be used as approximation of top-$k$ largest coordinates of gradient vector $\g$. 

Due to its linearity, the Count Sketch is widely adopted in distributed systems. Consider the case of a parameter server and two workers hosting vectors $\g_1$ and $\g_2$. To reduce communication, both workers can send the parameter server sketches $S(\g_1)$ and $S(\g_2)$ instead of the vectors themselves. The parameter server can then merge these sketches as $S(\g) = S(\g_1 + \g_2) = S(\g_1) + S(\g_2)$.
This lets the parameter server find the approximate top-$k$ largest coordinates in a vector distributed among many workers. We defer a more detailed discussion of the Count Sketch to Appendix~\ref{appendix:sketching}.

\section{Sketched SGD}

In \ssgd, each worker transmits a sketch of its gradient instead of the gradient itself, as described above. The parameter server sums the workers' sketches, and then recovers the largest gradient elements by magnitude from the summed sketch. To improve the compression properties of sketching, we then perform a second round of communication, in which the parameter server requests the exact values of the top-$k$, and uses the sum of those in the weight update. This algorithm for recovering top-$k$ elements from a sketch is summarized in Algorithm~\ref{alg:heavymix}. 

Every iteration, only $k$ values of each worker's gradient are included in the final weight update. Instead of discarding the remaining $d-k$ gradient elements, it is important both theoretically and empirically to accumulate these elements in local error accumulation vectors, which are then added to the next iteration's gradient \citep{karimireddy2019error,stich2018sparsified}. This process is summarized in Algorithm \ref{alg:sketchedtopkofsumsgd}.

\begin{algorithm}
\caption{\textsc{\hmx}}
\label{alg:heavymix}
\begin{algorithmic}[1]
	\small
	\REQUIRE $\bS$ - sketch of gradient $\sg$; $k$ - parameter
	\STATE Query $\hat \ell^2_2 = (1 \pm 0.5) \|\sg\|^2_2$ from sketch $S$
	\STATE $\forall i$ query $\hat \sg_i^2 = \sg_i^2 \pm \frac{1}{2k}\|\sg\|^2_2$ from sketch  $S$ 

	\STATE $H \leftarrow \left\{i | \hat \sg_i \ge \hat \ell_2^2/k \right\}$ and  $NH \leftarrow \left\{i|\;\hat \sg_i < \hat \ell_2^2/k \right\}$ \label{codeline:hmx2}
	
	\STATE $\text{Top}_k = H \cup \text{rand}_l(NH)$, where $l = k - |H|$ \label{codeline:hmx3}
	\STATE second round of communication to get exact values of Top$_k$
	\ENSURE $\tkg$: $\forall i \in \text{Top}_k: \tkg_i = \sg_i$ and  $\forall i \notin \text{Top}_k: \tkg_i = 0$
	\end{algorithmic}
\end{algorithm}

\begin{algorithm}
\caption{\textsc{sketched-sgd}}
\label{alg:sketchedtopkofsumsgd}
    \begin{algorithmic}[1]
    \small
	\REQUIRE $k,\xi, T, W$
		\STATE $\eta_t \leftarrow \frac{1}{t+\xi},q_t \leftarrow (\xi + t)^2, Q_T = \sum_{t=1}^T q_t, \ba_0=\mathrm{0}$
	\FOR{$t = 1,2,\cdots T$}
		\STATE Compute stochastic gradient $g^i_t$ \hfill$\text{Worker}_i$
	    \STATE Error correction: $\ag^i_t = \eta_t \sg^i_t + \ba^i_{t-1}$ \hfill $\text{Worker}_i$
		\STATE Compute sketches $\bS_t^i$ of $\ag_t^i$ and send to Parameter Server \hfill $\text{Worker}_i$
		\STATE Aggregate sketches $\bS_t = \frac{1}{W}\sum_{i=1}^W \bS_t^i$ \hfill Parameter Server 
	   	\STATE $\tkg_t$ = \textsc{\hmx}($\bS_t, k$) \hfill Parameter Server 
	    \STATE Update $ \w_{t+1} = \w_{t} - \tilde \g_{t}$ and send $\tilde \g_{t}$ (which is $k$-sparse) to Workers \hfill Parameter Server
      	\STATE Error accumulation: $\ba^i_t = \ag^i_t - \tkg_t$ \hfill $\text{Worker}_i$
     \ENDFOR 
	\ENSURE  $\hat \w_T = \frac{1}{Q_T} \sum_{t=1}^T q_t\w_t$
	\end{algorithmic}
\end{algorithm}

We now state convergence results for \ssgd. Proofs are deferred to Appendix~\ref{sec:proof}.
\begin{theorem}[strongly convex, smooth]
\label{thm:stronglyconvexsmooth}
Let $f:\R^d \rightarrow \R$ be a $L$-smooth $\mu$-strongly convex function, and let the data be shared among $W$ workers. Given $0<k\leq d, 0<\alpha, and \delta<1$, Algorithm~\ref{alg:sketchedtopkofsumsgd} \textsc{Sketched-SGD} run with sketch size $=\bigO{k\log(dT/\delta}$, step size $\eta_t =\frac{1}{t+\xi}$, with $\xi > 2 + \frac{d(1+\beta)}{k(1+\rho)}$, with $\beta>4$ and $\rho = \frac{4 \beta}{(\beta-4)(\beta+1)^2}$ after $T$ steps outputs $\hat \w_T$ such that the following holds,
\begin{CompactEnumerate}
    \item  With probability at least $1-\delta, \ \E{f(\hat \w_T)} - f(\w^*) \leq \bigO{\frac{\sigma^2}{ \mu T} + \frac{d^2G^2L}{k^2\mu^2 T^2}  + \frac{d^3G^3}{k^3\mu  T^3}}$
    \item The total communication per update is $\Theta(k \log(dT/\delta)W)$ bits.
\end{CompactEnumerate}
\end{theorem}

\paragraph{Remarks}
\begin{CompactEnumerate}
    \item The convergence rate for vanilla SGD is $\mathcal{O}(1/T)$. Therefore, our error is larger the SGD error when $T=o((d/k)^2)$, and approaches the SGD error for $T=\Omega((d/k)^2)$.
    \item Although not stated in this theorem, \citet{stich2018sparsified} show that using the top-$k$ coordinates of the true mini-batch gradient as the SGD update step yields a convergence rate equivalent to that of \ssgd. We therefore use this ``true top-$k$'' method as a baseline for our results.
    \item Note that the leading term in the error is $O(\sigma^2/T)$ (as opposed to $O(G^2/T)$ in \citep{stich2018sparsified}); this implies that in setting where the largest minibatch size allowed is too large to fit in one machine, and going distributed allows us to use larger mini-batches, the variance reduces by a factor $W$. This reduces the number of iterations required (asymptotically) linearly with $W$.
    \item As is standard, the above high probability bound can be converted to an expectation (over randomness in sketching) bound; this is stated as Theorem \ref{thm:sketchedsgdexp} in the Appendix~\ref{sec:proof}.
    \item The result of \citep{karimireddy2019error} allows us to extend our theorems to smooth nonconvex and non-smooth convex functions; these are presented as Theorems \ref{thm:ncs} and \ref{thm:cns} in the Appendix~\ref{sec:proof2}..
\end{CompactEnumerate}

\paragraph{Proof Sketch.} The proof consists of two parts. First, we show that \ssgd satisfies the criteria in \cite{stich2018sparsified}, from which we obtain a convergence result when running \ssgd on a single machine. We then use properties of the Count Sketch to extend this result to the distributed setting.

For the first part, the key idea is to show that our heavy hitter recovery routine \hmx \ satisfies a \emph{contraction} property, defined below.

\begin{definition}[$\tau$-contraction \citep{stich2018sparsified}]
A $\tau$-contraction operator is a possibly randomized operator $\text{comp}: \bbR^d \rightarrow \bbR^d$ that satisfies: $\forall \x \in \bbR^d, \ \E{\norm{\x - \text{comp}(\x)}^2} \leq \br{1 - \tau}\norm{\x}^2$
\end{definition}

Given a contraction operator with $\tau = k/d$, and assuming that the stochastic gradients $\sg$ are unbiased and bounded as $\E{\norm{\sg}^2} \leq G^2$,
choosing the step-size appropriately, \citet{stich2018sparsified} give  a convergence rate of $\bigO{\frac{G^2}{\mu T} + \frac{d^2G^2L}{k^2\mu^2T^2} + \frac{d^3G^3}{k^3\mu T^3}}$ for sparsified SGD with error accumulation. As stated in Lemma \ref{lem:contraction}, \hmx~ satisfies this contraction property, and therefore inherits this (single-machine) convergence result:

\begin{lemma}
\label{lem:contraction}
\hmx, with sketch size $\Theta(k \log (d/\delta))$ is a $k/d$-contraction with probability $\geq 1-\delta$.
\end{lemma}
This completes the first part of the proof. To extend \ssgd to the distributed setting, we exploit the fact that Count Sketches are linear, and can approximate $\ell_2$ norms.
 The full proof is deferred to Appendix \ref{sec:proof}.

\section{Empirical Results}
\subsection{Training Algorithm}
In practice, we modify \ssgd in the following ways
\begin{CompactItemize}
    \item We employ momentum when training. Following \citet{lin2017deep}, we use momentum correction and momentum factor masking. Momentum factor masking mitigates the effects of stale momentum, and momentum correction is a way to do error feedback on SGD with momentum \citep{karimireddy2019error}.
    \item We use the Count Sketch to identify heavy coordinates, however we perform an additional round of communication to collect the exact values of those coordinates. In addition, to identify the top $k$ heavy coordinates, we query the Count Sketch, and then each of the workers, for the top $Pk$ elements instead; this is a common technique used with sketching to improve stability. The total resulting communication cost is $Pk+|S|+k$ per worker, where $|S|$ is the size of the sketch, and the last $k$ corresponds to the the updated model parameters the parameter server must send back to the workers.
    \item We transmit gradients of the bias terms uncompressed. The number of bias terms in our models is $<1\%$ of the total number of parameters.
\end{CompactItemize}

Our emperical training procedure is summarized in Algorithm~\ref{alg:empirical_training}.

\begin{algorithm}
\caption{\textsc{Empirical Training}}
\label{alg:empirical_training}
\begin{algorithmic}[1]
	\small
	\REQUIRE $k, \eta_t, m, T$
    \STATE $\forall i: \mathrm{u}^{i}, \mathrm{v}^{i} \leftarrow 0$
    \STATE Initialize $w_0^i$ from the same random seed on each Worker.
	\FOR{$t = 1,2,\ldots T$}
    	\STATE Compute stochastic gradient $\mathrm{g}^{i}_t$  \hfill$\text{Worker}_i$
    	\STATE Momentum: $\mathrm{u}^{i} \leftarrow m \mathrm{u}^{i} +  \mathrm{\sg}^{i}_t$  \hfill$\text{Worker}_i$
    	\STATE Error accumulation: $\mathrm{v}^{i} \leftarrow \mathrm{v}^{i} + \mathrm{u}^{i}$  \hfill$\text{Worker}_i$
    	\STATE Compute sketch $\bS^{i}_t$ of $\mathrm{v}^{i}$ and send to Parameter Server \hfill$\text{Worker}_i$
		\STATE Aggregate sketches $\bS_t = \frac{1}{W}\sum_{i=1}^W \bS_t^i$  \hfill Parameter Server 
        \STATE Recover the top-$Pk$ coordinates from $S_t$: $\tkg_t = top_{Pk}(S_t)$ \hfill Parameter Server 
        \STATE Query all workers for exact values of nonzero elements in $\tkg_t$; store the sum in $\tkg_t$ \hfill Parameter Server
        \STATE Send the $k$-sparse $\tkg_t$ to Workers \hfill Parameter Server
        \STATE update $\w_{t+1}^i=\w_t^i-\eta_t \tkg_t$ on each worker \hfill $\text{Worker}_i$
  	    \STATE $\mathrm{u}^i, \mathrm{v}^{i} \leftarrow 0,  \text{ for all } i \text{ s.t. } \tilde{\mathrm{g}}^i_t \neq 0$  \hfill$\text{Worker}_i$
     \ENDFOR 
	\end{algorithmic}
\end{algorithm}
\vspace{-10 pt}

\begin{figure}
    \centering
    \includegraphics[width=0.70\textwidth]{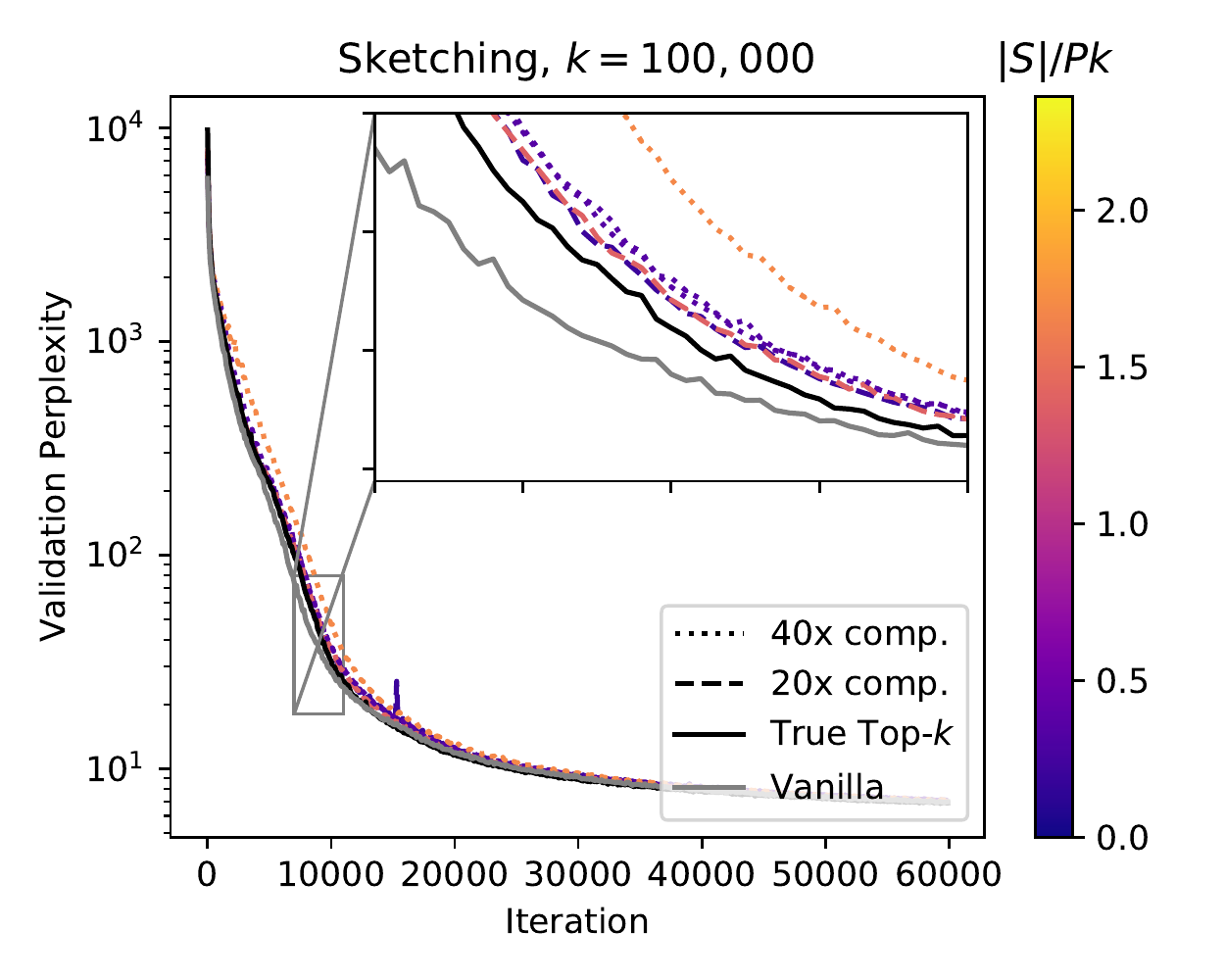}
    \caption{Learning curves for a transformer model trained on the WMT 2014 English to German translation task. All models included here achieve comparable BLEU scores after 60,000 iterations (see Table \ref{tab:bleu}). Each run used 4 workers.}
    \label{fig:sketched_lc}
\end{figure}

\subsection{Sketching Implementation}
We implement a parallelized Count Sketch with PyTorch \citep{pytorch}. The Count Sketch data structure supports a query method, which returns a provable $\pm \varepsilon \|\g \|_2$ approximation to each coordinate value.
However, to the best of our knowledge, there is no efficient way to find heavy coordinates in the presence of negative inputs. Fortunately, in our application, it is computationally efficient on the GPU to simply query the sketch for every gradient coordinate, and then choose the largest elements.

\subsection{Large \texorpdfstring{$d$}{\emph{d}}}
\label{sec:transformer}
First, we show that \ssgd achieves high compression with no loss in accuracy. Because the sketch size grows as $\mathcal{O}(\log d)$, we expect to see the greatest compression rates for large $d$. Accordingly, we test on a transformer model with ~90M parameters, and on a stacked LSTM model with ~73M parameters. We train both models on the WMT 2014 English to German translation task, and we use code from the OpenNMT project \citep{opennmt}. In all cases, the compression factor for \ssgd is computed as $2d/(|S|+Pk+k)$, where $2d$ is the cost to send a (dense) gradient and receive a new (dense) parameter vector, $|S|$ is the sketch size, $Pk$ is the number of elements sent in the second round of communication, and the last $k$ represents the number of modified parameter values that must be sent back to each worker.

\begin{figure}
    \centering
    \subfloat[WMT14 Translation Task]{\includegraphics[width=0.5\textwidth]{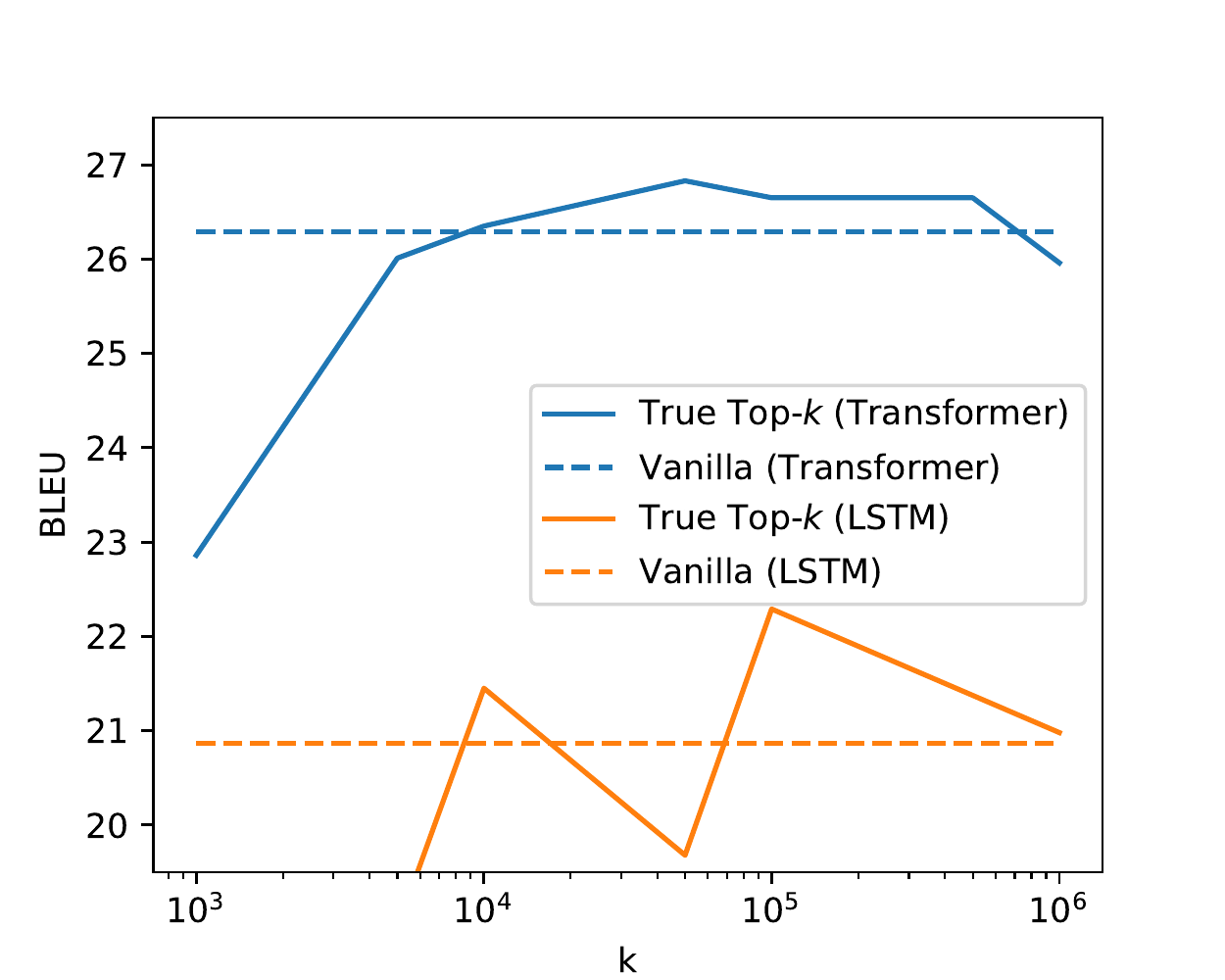}}
    \subfloat[CIFAR-10 Classification Task]{\includegraphics[width=0.5\textwidth]{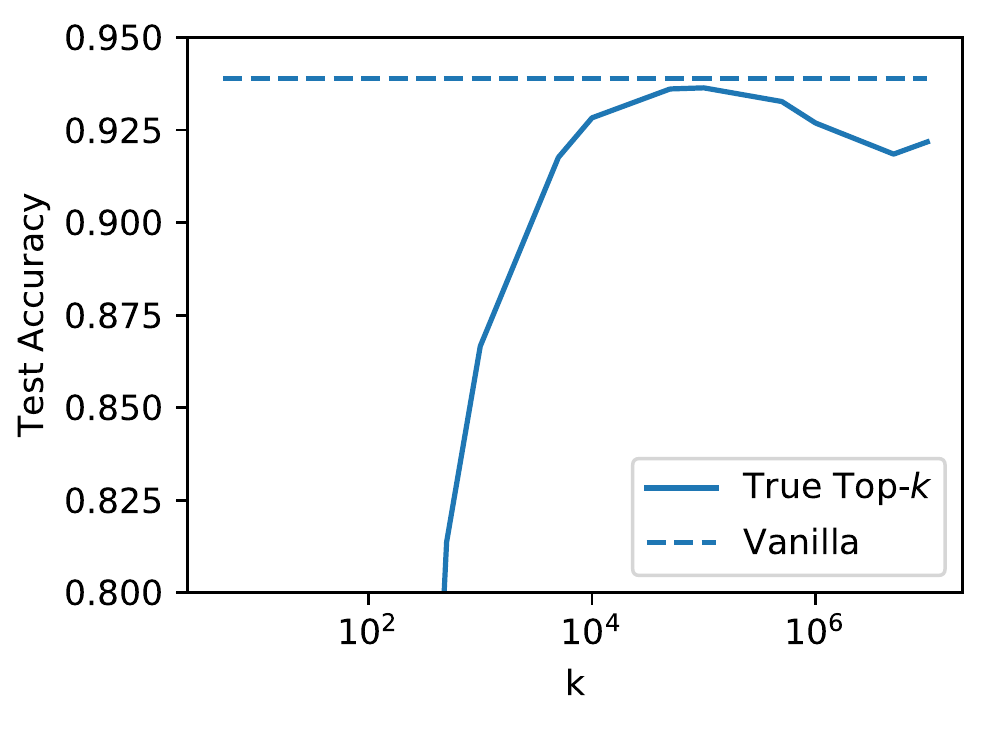}}
    \caption{True top-$k$ results for a range of $k$. Left: two models (transformer and LSTM) on the WMT 2014 English to German translation task. Right: a residual network on the CIFAR-10 classification task. For the larger models (left), true top-$k$ slightly outperforms the baseline for a range of $k$. We suspect this is because $k$-sparsifying gradients serves to regularize the model.}
    \label{fig:topk}
\end{figure}

\ssgd achieves the same theoretical convergence rate as top-$k$ SGD, in which the weight update consists of the top-$k$ elements of the full mini-batch gradient.
We therefore perform experiments with \ssgd using a value of $k$ that yields good performance for top-$k$ SGD.
Figure \ref{fig:topk} shows top-$k$ results over a range of values of $k$.
Curiously, performance starts to degrade for large $k$.
Although performance on the training data should in principle strictly improve for larger $k$, sparsifying gradients regularizes the model, so $k<d$ may yield optimal performance on the test set.
In addition, we expect performance to degrade on both the training and test sets for large $k$ due to momentum factor masking.
To mitigate stale momentum updates, momentum factor masking zeros the velocity vector at the $k$ coordinates that were updated in each iteration. In the limit $k=d$, this completely negates the momentum, hindering convergence.
For all \ssgd experiments on these two models, we use $k=100,000$, for which top-$k$ SGD yields a BLEU score of 26.65 for the transformer and 22.2 for the LSTM. For reference, uncompressed distributed SGD with the same hyperparameters achieves a BLEU of 26.29 for the transformer and 20.87 for the LSTM. Using \ssgd, we can obtain, with no loss in BLEU, a 40x reduction in the total communication cost during training, including the cost to disseminate updated model parameters. See Table \ref{tab:bleu} for a summary of BLEU results. Compression numbers include both the communication required to send gradients as well as the cost to send back the new model parameters. We do not include the cost to request the $Pk$ coordinates, nor to specify which $k$ model parameters have been updated, since these quantities can be efficiently coded, and contribute little to the overall communication.

Given that our algorithm involves a second round of communication in which $Pk$ gradient elements are transmitted, we investigate the tradeoff between a large sketch size and a large value of $P$. Approaching a sketch size of zero corresponds to using a weight update that is the top-$k$ of a randomly chosen set of $Pk$ gradient coordinates. Experiments with extremely small sketch size $|S|$ or extremely small values of $P$ tended to diverge or achieve very low BLEU score. For values of $|S|/Pk$ closer to $1$, we plot learning curves in Figure \ref{fig:sketched_lc}. As expected, uncompressed SGD trains fastest, followed by top-$k$ SGD, then 20x compression \ssgd, then 40x compression \ssgd. For the two 20x compression runs, the ratio of the sketch size to the number of exact gradient values computed has little effect on convergence speed. However, the higher compression runs prefer a relatively larger value of $P$.

\begin{table}
    \centering
    \begin{tabular}{r|cc}
         & BLEU (transformer) & BLEU (LSTM) \\
         Uncompressed Distributed SGD & 26.29 & 20.87 \\
         Top-$100,000$ SGD & 26.65 & 22.2\\
         \ssgd, 20x compression & 26.87\footnote{Sketch size: 5 rows by 1M columns; $P=36$.} & -- \\
         \ssgd, 40x compression & 26.79\footnote{Sketch size: 15 rows by 180,000 columns; $P=16$.} & 20.95 \footnote{Sketch size: 5 rows by 180,000 columns, $P=26$} \\
    \end{tabular}
    \vspace{10px}
    \caption{BLEU scores on the test data achieved for uncompressed distributed SGD, top-$k$ SGD, and \ssgd with 20x and 40x compression. Compression rates represent the total reduction in communication, including the cost to transmit the updated model parameters. Larger BLEU score is better. For both models, top-$k$ SGD with $k=100,000$ achieves a higher BLEU score than uncompressed distributed SGD. This difference may be within the error bars, but if not, it may be that stepping in only the direction of the top-$k$ is serving as a regularizer on the optimizer. Our main experiments are on the transformer model, for which we run additional experiments using 20x compression that we did not complete for the LSTM model.}
    \label{tab:bleu}
\end{table}

\subsection{Large \texorpdfstring{$W$}{\emph{W}}}
\label{sec:cifar}

\begin{figure}[t]
    \centering
    \includegraphics[width=0.75\textwidth]{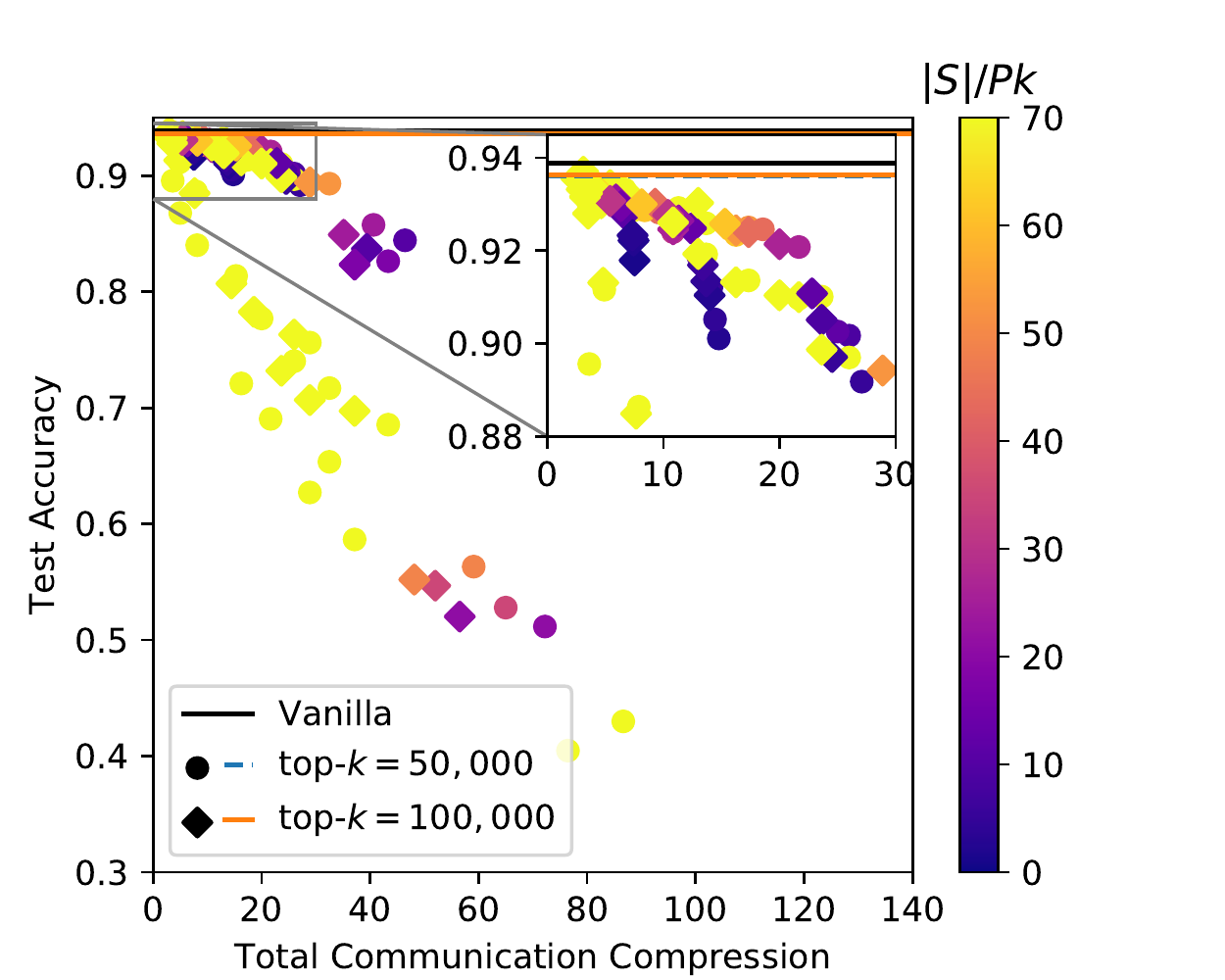}
    \caption{Tradeoff between compression and model accuracy for a residual network trained on CIFAR-10. We show results for $k=50,000$ as well as $k=100,000$, and color code each trained model based on the ratio of sketch size to the cost of the second round of communication. The (nearly overlapping) solid orange and dashed blue lines show the accuracy achieved by top$-k$ SGD for the two values of $k$, and the black line shows the accuracy achieved by uncompressed distributed SGD. All models in this plot were trained with 4 workers.}
    \label{fig:c_compression}
\end{figure}

\begin{figure}
    \centering
    \includegraphics[width=0.65\textwidth]{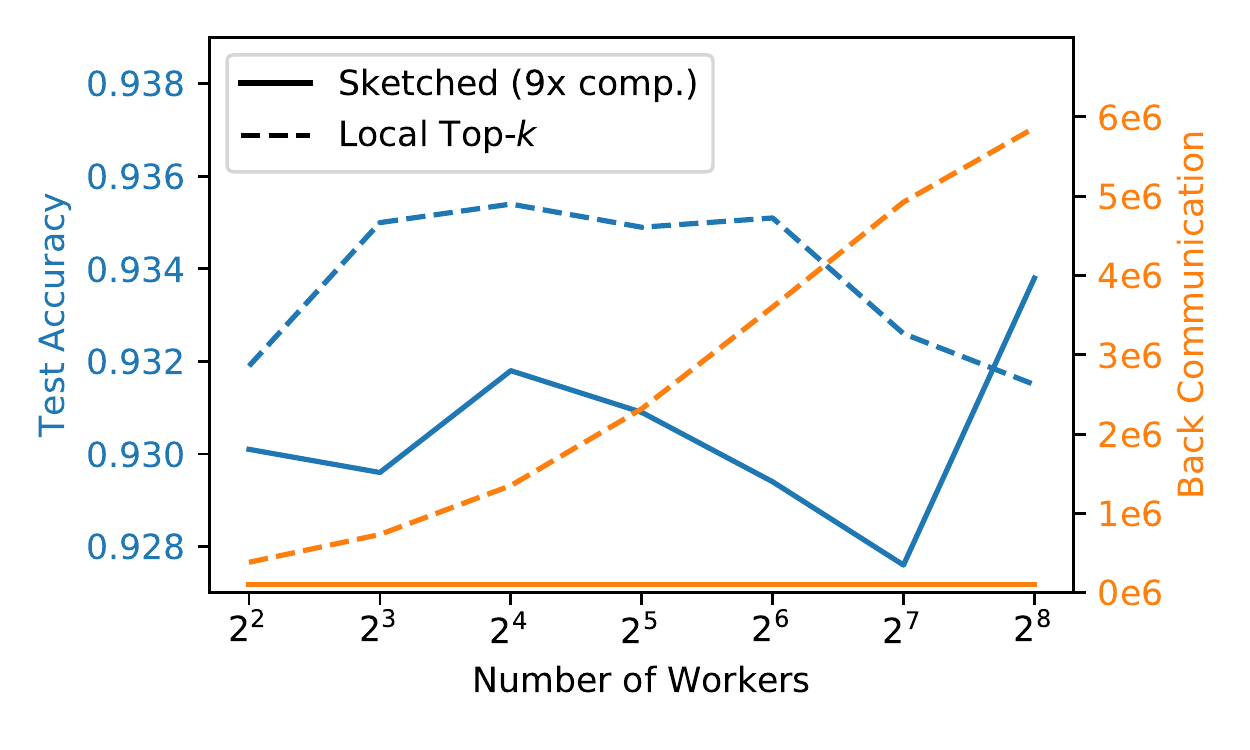}
    \caption{Comparison between \ssgd and local top-$k$ SGD on CIFAR10. Neither algorithm sees an appreciable drop in performance for up to 256 workers, but the amount of communication required for local top-$k$ grows quickly to $\approx d=6.5\times 10^6$ as the number of workers increases. As a result, the best overall compression that local top-$k$ can achieve for many workers is 2x.}
    \label{fig:c_num_workers}
\end{figure}

To re-iterate, the per-worker communication cost for \ssgd is not only sub-linear in $d$, but also independent of $W$.
To demonstrate the power of this experimentally, we train a residual network on the CIFAR-10 dataset with \ssgd, using up to 256 workers \citep{cifar10}.
We compare to local top-$k$, a method where each worker computes and transmits only the top-$k$ elements of its gradient.
The version of local top-$k$ SGD we compare to is similar to Deep Gradient Compression, except we do not clip gradients, and we warm up the learning rate instead of the sparsity \citep{lin2017deep}.
Results are shown in Figure~\ref{fig:c_num_workers}.
Neither algorithm sees an appreciable drop in accuracy with more workers, up to $W=256$.
However, while the communication cost of \ssgd is constant in $W$, the communication cost for local top-$k$ scales with $W$ until reaching $\Theta(d)$.
This scaling occurs because the local top-$k$ of each worker might be disjoint, leading to as many as $kW$ parameters being updated.
In practice, we do in fact observe nearly linear scaling of the number of parameters updated each iteration, until saturating at $d$ (dashed orange line in Figure~\ref{fig:c_num_workers}).
For $W=256$, the communication of the updated model parameters back to each worker is nearly dense ($d\approx 6.5\times 10^6$), reducing the overall compression of local top-$k$ to at best $\sim 2\times$.

For a fixed small number of workers ($W=4$), we also investigate the tradeoff between compression rate and final test accuracy.
Figure \ref{fig:c_compression} shows this tradeoff for two values of $k$ and a wide range of sketch sizes and values of $P$.
As expected, increasing the compression rate leads to decreasing test accuracy.
In addition, as evidenced by the color coding, using a very large sketch size compared to $Pk$ tends to yield poor results.
Although high compression rates decrease accuracy, in our experience, it is possible to make up for this accuracy drop by training longer.
For example, choosing one of the points in Figure \ref{fig:c_compression}, training with 17x compression for the usual number of iterations gives 92.5\% test accuracy.
Training with 50\% more iterations (reducing to 11x overall compression) restores accuracy to 94\%.
In Figure \ref{fig:c_compression}, every model is trained for the same number of iterations.

\section{Discussion}
In this work we introduce \ssgd, an algorithm for reducing the communication cost in distributed SGD using sketching. We provide theoretical and experimental evidence that  our method can help alleviate the difficulties of scaling SGD to many workers. While uncompressed distributed SGD requires communication of size $2d$, and other gradient compressions improve this to $\mathcal{O}(d)$ or $\mathcal{O}(W)$, \ssgd further reduces the necessary communication to $\mathcal{O}(\log d)$. Besides reducing communication, our method provably converges at the same rate as SGD, and in practice we are able to reduce the total communication needed by up to 40x without experiencing a loss in model quality.

A number of other techniques for efficient training could be combined with \ssgd, including gradient quantization and asynchronous updates. We expect that the advantages asynchronous updates bring to regular SGD will carry over to \ssgd. And given that elements of gradient sketches are sums of gradient elements, we expect that quantizing sketches will lead to similar tradeoffs as quantizing the gradients themselves. Preliminary experiments show that quantizing sketches to 16 bits when training our ResNets on CIFAR-10 leads to no drop in accuracy, but we leave a full evaluation of combining quantization, as well as asynchronous updates, with \ssgd to future work.

Machine learning models are constantly growing in size (e.g. OpenAI's GPT-2, a transformer with 1.5 billion parameters \citep{gpt2}), and training is being carried out on a larger and larger number of compute nodes. As communication increasingly becomes a bottleneck for large-scale training, we argue that a method that requires only $\mathcal{O}(\log d)$ communication has the potential to enable a wide range of machine learning workloads that are currently infeasible, from highly parallel training in the cloud, to Federated Learning at the edge \citep{federated1}.

\section{Acknowledgements}
This research was supported, in part, by NSF BIGDATA grants IIS-1546482 and IIS-1838139, NSF CAREER grant 1652257, ONR Award N00014-18-1-2364 and the Lifelong Learning Machines program from DARPA/MTO.
This material is based upon work supported by the National Science Foundation Graduate Research Fellowship under Grant No. DGE 1752814.

\bibliographystyle{plainnat}
\bibliography{main}

\newpage

\appendix
\newpage
\begin{center}
\textbf{ \Large Supplementary}
\end{center}
\section{Proofs} 
\label{sec:proof}

\begin{proof}[Proof of Lemma \ref{lem:contraction}]
Given $\sg \in \bbR$, the \hmx~algorithm extracts all ${(1/k, \ell^2_2)\text{-heavy}}$ elements from a Count Sketch $S$ of $\sg$. 
Let $\hat \sg$ be the values of all elements recovered from its sketch. For a fixed $k$, we create two sets $H$ (heavy), and $NH$ (not-heavy). All coordinates of $\hat \sg$ with values at least $\frac{1}{k}  \hat \ell_2^2$ are put in $H$, and all others in $NH$, where $\hat \ell_2$ is the estimate of $\|\sg\|_2$ from the Count Sketch. 
Note that the number of elements in $H$ can be at most $k$. 
Then, we sample uniformly at random $l = k - |H|$ elements from $NH$ , and finally output its union with $H$. 
We then do a second round of communication to get exact values of these $k$ elements.

Note that, because of the second round of communication in \hmx~and the properties of the Count Sketch, with probability at least $1-\delta$ we get the exact values of all elements in $H$. Call this the ``heavy hitters recovery'' event. 
Let $\g_H$ be a vector equal to $\g$ at the coordinates in $H$, and zero otherwise. Define $\g_{NH}$ analogously.
Conditioning on the heavy hitters recovery event, and taking expectation over the random sampling, we have
\begin{align*}
    \E{\norm{\g - \tkg}^2} &= \norm{\g_H - \bar\g_H}^2 + \E{\norm{\g_{NH} - \text{rand}_{l}\br{\g_{NH}}}^2} \\
    & \leq \br{1 - \frac{k -\abs{H}}{d-\abs{H}}}\norm{\g_{NH}}^2  \leq \br{1 - \frac{k -\abs{H}}{d-\abs{H}}} \br{1 - \frac{\abs{H}}{2k}}\norm{\g}^2
\end{align*}
Note that, because we condition on the heavy hitter recovery event, $\bar \g_H = \g_H$ due to the second round communication 
(line 9 of Algorithm \ref{alg:empirical_training}). The first inequality follows using Lemma 1 from \citet{stich2018sparsified}. The second inequality follows from the fact that the heavy elements have values at least $\frac{1}{k} \hat \ell_2^2 \geq \frac{1}{2k}\norm{\g}^2$, and therefore $\norm{\g_{NH}}^2 = \norm{\g}^2 - \norm{\g_H}^2 \leq \br{1 - \frac{\abs{H}}{2k}}\norm{\g}^2.$

Simplifying the expression, we get 
\begin{align*}
    \E{\norm{\g - \tkg}^2} 
      & \leq \br{\frac{2k - \abs{H}}{2k}} \br{\frac{d-k}{d-\abs{H}}} \norm{\g}^2  = \br{\frac{2k - \abs{H}}{2k}}\br{\frac{d}{d-\abs{H}}} \br{1 - \frac{k}{d}} \norm{\g}^2.
\end{align*}
Note that the first two terms can be bounded as follows:
\small
\begin{align*}
\br{\frac{2k  - \abs{H}}{2k}}\br{\frac{d}{d-\abs{H}}} \leq 1 & 
\iff kd -\abs{H}d \leq kd - 2k \abs{H} \iff \abs{H}(d - 2k) \geq 0 
\end{align*}
\normalsize
which holds when $k \leq d/2$ thereby completing the proof.

\end{proof}

\subsection{Proof of the main theorem}
\begin{proof}[Proof of Theorem \ref{thm:stronglyconvexsmooth}]
First note that, from linearity of sketches), the top-$k$ (or heavy) elements from the merged sketch $\bS_t = \sum_{i=1}^W \bS^i_t$ are the top-$k$ of the sum of vectors that were sketched. 
We have already shown in Lemma \ref{lem:contraction} that that extracting the top-$k$ elements from $\bS-T$ using \hmx~gives us a $k$-contraction on the sum of gradients.
Moreover since the guarantee is relative and norms are positive homogeneous, the same holds for the average, i.e. when dividing by $W$.
Now since the average of stochastic gradients is still an unbiased estimate, this reduces to \textsc{Sketched-SGD} on one machine, and the convergence therefore follows from Theorem \ref{thm:sketchedsgd2}.
\end{proof}

A key ingredient is the result in the one machine setting, stated below.
\begin{theorem}
\label{thm:sketchedsgd2}
Let $f:\R^d \rightarrow \R$ be a $L$-smooth $\mu$-strongly convex function. Given $T>0$ and $0<k\leq d,0<\delta<1$, and a $\tau_k$-contraction, Algorithm \ref{alg:sketchedtopkofsumsgd} \textsc{Sketched-SGD} with sketch size $\bigO{k \log(dT/\delta)}$ and step size $\eta_t =\frac{1}{t+\xi}$, with $\xi > 1 + \frac{1+\beta}{\tau_k(1+\rho)}$, with $\beta>4$ and $\rho = \frac{4 \beta}{(\beta-4)(\beta+1)^2}$, after $T$ steps outputs $\hat \w_T$ such that with probability at least $1-\delta$
\small
\begin{align*}
    \E{f(\hat \w_T)} - f(\w^*)  &\leq \bigO{\frac{\sigma^2}{\mu T} + \frac{G^2L}{\tau_k^2\mu^2T^2} + \frac{G^3}{\tau_k^3\mu T^3}}
\end{align*}
\normalsize
\end{theorem}
\begin{proof}[Proof of Theorem \ref{thm:sketchedsgd2}]
The proof, as in \cite{stich2018sparsified}, just follows using convexity and Lemmas \ref{lem:accumulation},\ref{lem:perturbation} and fact \ref{lem:recurrence}.
The lemmas which are exactly same as \cite{stich2018sparsified}, are stated as facts. However, the proofs of lemmas, which change are stated in full for completeness, with the changes highlighted.

From convexity we have that 
\begin{align*}
    f\br{\frac{1}{Q_T}\sum_{i=1}^T q_t\w_t} - f(\w^*) & \leq \frac{1}{Q_T} \sum_{t=1}^T q_t f(\w_t) - f(\w^*) = \frac{1}{Q_T} \sum_{t=1}^T  q_t  \br{ f(\w_t) - f(\w^*)}
\end{align*}
Define $\epsilon_t =  f(\w_t) - f(\w^*)$, the excess error of iterate $t$. From Lemma \ref{lem:perturbation} we have,
\begin{align*}
    \E{\norm{\tilde \w_{t+1}-\w^*}^2} &\leq \br{1-\frac{\eta_t \mu}{2}} \E{\norm{\tilde \w_{t}-\w^*}^2} +\sigma^2\eta_t^2  -  \br{1-\frac{2}{\xi}}\epsilon_t \eta_t+ (\mu + 2L)\E{\norm{\ba_t}^2}\eta_t
\end{align*}

Bounding the last term using Lemma \ref{lem:accumulation}, with probability at least $1 -\delta$, we get,

\begin{align*}
   &\E{\norm{\tilde \w_{t+1}-\w^*}^2} \leq \br{1-\frac{\eta_t \mu}{2}} \E{\norm{\tilde \w_{t}-\w^*}^2} +\sigma^2 \eta_t^2  - \br{1-\frac{2}{\xi}} \epsilon_t \eta_t + \frac{(\mu + 2L)4\beta  G^2 }{\tau_k^2(\beta-4)}\eta_t^3
\end{align*}

where $\tau_k$ is the contraction we get from \hmx. We have alreay show that $\tau_k \leq \frac{k}{d}$. 

Now using Lemma \ref{lem:recurrence} and the fist equation, we get,
\begin{align*}
    &f\br{\frac{1}{Q_T}\sum_{i=1}^T q_t\w_t} - f(\w^*)  \leq \frac{\mu \xi^4 \E{\norm{\w_{0}-\w^*}^2}}{8(\xi-2) Q_T} + \frac{4T(T+2\xi)\xi\sigma^2}{\mu (\xi-2)Q_T} + \frac{256(\mu + 2L)\beta \xi G^2 T}{\mu^2 (\beta-4)\tau_k^2 (\xi-2)Q_T}
\end{align*}

Note that $\xi> 2+ \frac{1+\beta}{\tau_k(1+\rho)}$. Moreover $Q_T = \sum_{t=1}^T q_t = \sum_{t=1}^T (\xi+t)^2 \geq \frac{1}{3}T^3$ upon expanding and using the conditions on $\xi$. Also $\xi/(\xi -2) = \bigO{1 + 1/\tau_k}$.

Finally using $\sigma^2\leq G^2$ and Fact \ref{lem:rakhlin} to bound $\E{\norm{\w_{0}-\w^*}^2} \leq 4G^2/\mu^2$ completes the proof.

\end{proof}
\begin{lemma}
\label{lem:perturbation}
Let $f:\R^d \rightarrow \R$ be a $L$-smooth $\mu$-strongly convex function, and $\w^*$ be its minima. Let $\bc{ \w_t}_t$ be a sequence of iterates generated by Algorithm \ref{alg:sketchedtopkofsumsgd}.
 
Define error $\epsilon_t: = \E{f(\w_t) - f(\w^*)}$ and $\tilde \w_{t+1} = \tilde \w_{t} - \eta_{t}\sg_t$ be a stochastic gradient update step at time $t$, with $\E{\norm{\sg_t - \nabla f(\w_t)}^2}\leq \sigma^2$, $\E{\norm{\sg_t}^2}\leq G^2$ and $\eta_t = \frac{1}{\mu(t+\xi)}, \xi > 2$ then we have,
\begin{align*}
    \E{\norm{\tilde \w_{t+1}-\w^*}^2} &\leq \br{1-\frac{\eta_t \mu}{2}} \E{\norm{\tilde \w_{t}-\w^*}^2} +\sigma^2 \eta_t  -  \br{1-\frac{2}{\xi}}\epsilon_t \eta_t+ (\mu + 2L)\E{\norm{\ba_t}^2}\eta_t
\end{align*}
\end{lemma}
\begin{proof}

This is the first step of the perturbed iterate analysis framework \cite{mania2015perturbed}. We follow the steps as in  \cite{stich2018sparsified}. The only change is that the proof of \cite{stich2018sparsified} works with bounded gradients i.e.  $\E{\norm{\sg}^2} \leq G^2$. This assumption alone, doesn't provide the variance reduction effect in the distributed setting. We therefore adapt the analysis with the the variance bound $\E{\norm{\sg - \nabla f(\w)}^2} \leq \sigma^2$.
 \begin{align*}
    &\norm{\tilde \w_{t+1} - \w^*}^2 =\norm{\tilde \w_{t+1} - \tilde \w_t + \tilde \w_t -\w^*}^2 =  \norm{\tilde \w_{t+1} - \tilde \w_t}^2 + \norm{\tilde \w_t - \w^*}^2 + 2\ip{\tilde \w_{t} - \w^*}{\tilde \w_{t+1}-\tilde \w_t} \\
     &= \eta_t^2\norm{g_t}^2 + \norm{\tilde \w_t - \w^*}^2 + 2\ip{\tilde \w_{t} - \w^*}{\tilde \w_{t+1}-\tilde \w_t} = \eta_t^2\norm{g_t-\nabla f(\w_t)}^2 + \eta_t^2 \norm{\nabla f(\w_t)}^2 +  \norm{\tilde \w_t - \w^*}^2\\ &+ 2 \eta_t\ip{\sg_t - \nabla f(\w_t)}{\nabla f(\w_t)}  + 2\eta_t\ip{\w^* - \tilde \w_{t}}{\sg_t}
\end{align*}

Taking expectation with respect to the randomness of the last stochastic gradient, we have that the term $\ip{\sg_t - \nabla f(\w_t)}{\nabla f(\w_t)}=0$ by $\E{\sg_t} = \nabla f(\w_t)$. Moreover, the term $\E{\sg_t - \nabla f(\w_t)}^2 \leq \sigma^2$. We expand the last term as,
\begin{align*}
    \ip{\w^*-\tilde \w_{t}}{\nabla f(\w_t)} = \ip{\w^* -\w_t}{\nabla f(\w_t)} +  \ip{\w_{t} - \tilde \w_t}{\nabla f(\w_t)}
\end{align*}
The first term is bounded by $\mu$-strong convexity as,
\begin{align*}
    &f(\w^*) \geq f(\w_t) + \ip{\nabla f(\w_t)}{\w^* - \w_t} + \frac{\mu}{2}\norm{ \w_t - \w^*}^2 \\
    & \iff \ip{\nabla f(\w_t)}{\w^* - \w_t} \leq f(\w^*) - f(\w_t) - \frac{\mu}{2}\norm{ \w_t - \w^*}^2 \\
    &\leq-\epsilon_t + \frac{\mu}{2}\mu\norm{\tilde \w_t - \w_t}  - \frac{\mu}{4} \norm{\w^* - \tilde \w_t}
\end{align*}
where in the last step, we define $\epsilon_t :=f(\w_t) - f(\w^*)$ and use $\norm{\bu + \bv}^2 \leq 2(\norm{\bu}^2 + \norm{\bv}^2)$. The second term is bounded by using $2\ip{\bu}{\bv} \leq a \norm{\bu}^2 + \frac{1}{a}\norm{\bv}^2$ as follows,
\begin{align*}
   2 \ip{\w_{t} - \tilde \w_{t}}{\nabla f(\w_t)} \leq 2L\norm{\w_{t} - \tilde \w_{t}}^2 + \frac{1}{2L}\norm{\nabla f(\w_t)}^2
\end{align*}
Moreover, from Facr \ref{lem:boundgrad}, we have $\norm{\nabla f(\w_t)}^2 \leq 2 L \epsilon_t$. Taking expectation and putting everything together, we get,
\begin{align*}
\E{\norm{\tilde \w_{t+1} - \w^*}^2}
&\leq \br{1-\frac{\mu \eta_t}{2}}\E{\norm{\tilde \w_t - \w^*}^2} + \eta_t^2 \sigma^2 \\&+ \br{\mu + 2L}\eta_t \E{\norm{\w_t - \tilde \w_t}^2} + \br{2L \eta_t^2 - \eta_t}\epsilon_t
\end{align*}
We now claim that the last term $2L \eta_t^2 - \eta_t \leq - \frac{\xi - 2}{\xi} \eta_t$ or equivalently $2L \eta_t^2 - \br{1 - \frac{\xi - 2}{\xi}}\eta_t \leq 0$. Note that this is a quadratic  in $\eta_t$ which is satisfied between its roots $0$ and $\frac{1}{L\xi}$. So it suffices to show is that our step sizes are in this range. In particular, the second root (which is positive by choice of $\xi$) should be no less than step size. We have $\eta_t = \frac{1}{\mu(t+\xi)}$, $\eta_t \leq \frac{1}{\mu \xi} \ \forall \ t$, the second root $\frac{1}{L\xi} \geq \frac{1}{\mu \xi}$ because smoothness parameter $L \geq \mu$, the strong convexity parameter, or equivalently the condition number $\kappa:=L/\mu \geq 1$.  
Combining the above with $\ba_t = \w_t-\tilde \w_t$,  we get,
\begin{align*}
\E{\norm{\tilde \w_{t+1} - \w^*}^2} 
&\leq \br{1-\frac{\mu \eta_t}{2}}\E{\norm{\tilde \w_t - \w^*}} + \eta_t^2 \sigma^2 \\&+ \br{\mu + 2L}\eta_t \E{\norm{\ba_t}^2} -\br{1-\frac{2}{\xi}} \eta_t\epsilon_t
\end{align*}

\end{proof}
\begin{fact} \cite{rakhlin2012making}
\label{lem:rakhlin}
Let $f:\R^d \rightarrow$ be a $\mu$-strongly convex function, and $\w^*$ be its minima. Let $\sg$ be an unbiased stochastic gradient at point $\w$ such that $\E{\norm{\sg}^2} \leq G^2$, then

\begin{align*}
    \E{\norm{ \w - \w^*}^2} \leq \frac{4G^2}{\mu^2}
\end{align*}
\end{fact}

\begin{fact}
\label{lem:boundgrad}
For $L$-smooth convex function $f$ with minima $\w^*$, then the following holds for all points $\w$, 
\begin{align*}
    \norm{\nabla f(\w) - \nabla  f(\w^*)}^2 \leq 2L (f(\w)-f(\w^*))
\end{align*}
\end{fact}

\begin{fact} \cite{stich2018sparsified}
\label{lem:recurrence}
Let $\bc{b_t}_{t\geq 0}, b_t \geq 0$ and $\bc{\epsilon_t}_{t\geq 0}, \epsilon_t \geq 0$ be sequences such that,
\begin{align*}
    b_{t+1} \leq \br{1-\frac{\mu \eta_t}{2}}b_t - \epsilon_t \eta_t + A \eta^2 + B \eta^3  
\end{align*}
for constants $A,B >0, \mu\geq 0, \xi > 1$. Then,
\begin{align*}
\frac{1}{Q_T}\sum_{t=0}^{T-1} q_t \epsilon_t \leq \frac{\mu \xi^3 b_0}{8 Q_T} + \frac{4T(T+2\xi)A}{\mu Q_T} + \frac{64TB}{\mu^2 Q_T}
\end{align*}
for $\eta_t = \frac{8}{\mu(\xi+t)}, q_t = (\xi + t)^2, Q_T = \sum_{t=0}^{T-1}{q_t} \geq \frac{T^3}{3}$
\end{fact}

\begin{fact} \cite{stich2018sparsified}
\label{lem:accumulation2}
Let $\bc{h_t}_{t>0}$ be a sequence satisfying $ h_0 = 0 $ and
\begin{align*}
    h_{t+1} \leq \min \bc{ \br{1 - \tau/2}h_{t} + \frac{2}{\tau_k}\eta_t^2 A, (t+1) \sum_{i=0}^{t}\eta_i^2 A}
\end{align*}
for constant $A>0$, then with $\eta_t = \frac{1}{t+\xi}$ with $\xi > 1 + \frac{1+\beta}{\tau_k(1+\rho)}$, with $\beta>4$ and $\rho = \frac{4 \beta}{(\beta-4)(\beta+1)^2}$, for $t\geq0$ we get,
\begin{align*}
h_t \leq \frac{4\beta}{(\beta-4)}\cdot \frac{\eta_t^2 A}{\tau_k^2}
\end{align*}

\end{fact}

\begin{lemma}
\label{lem:accumulation}
With probability at least $1-\delta$
\begin{align*}
\E{\norm{\ba_t}^2} \leq \frac{4\beta}{(\beta-4)}\cdot \frac{\eta_t^2 G^2}{\tau_k^2}
\end{align*}
\end{lemma}

\begin{proof}[Proof of Lemma \ref{lem:accumulation}]
The proof repeats the steps in \cite{stich2018sparsified} with minor modifications. In particular, the compression is provided by the recovery guarantees of Count Sketch, and we do a union bound over all its instances. We write the proof in full for the sake of completeness. Note that
\begin{align*}
    \ba_{t} = \ba_{t-1} + \eta_{t-1}\sg_{t-1} - \tkg_{t-1}
\end{align*}
We first claim that $\E{\norm{\ba_t}^2} \leq t \eta_t^2 G^2$.
Since $\ba_0 = 0$, we have $\ba_{t} = \sum_{i=1}^{t}(\ba_{i} - \ba_{i-1}) = \sum_{i=0}^{t-1} (\eta_i\sg_i - \tkg_i)$. Using $(\sum_{i=1}^n a_i)^2 \leq (n+1) \sum_{i=1}^n a_i^2$ and taking expectation, we have

\begin{align*}
    \E{\norm{\ba_t}^2}  \leq  t \sum_{i=0}^{t-1}\E{\norm{\eta_i\sg_i - \tkg_i}^2} \leq t \sum_{i=0}^{t-1}\eta_i^2 G^2
\end{align*}

Also, from the guarantee of Count Sketch, we have that, with probability at least $1-\delta/T$, the following holds give that our compression is a $\tau_k$ contraction.

Therefore
\begin{align*}
    \norm{\ba_{t+1}}^2  &\leq (1-\tau_k) \norm{\ba_{t}+ \eta_t\sg_t}^2 \\
\end{align*}
Using inequality $(a+b)^2 \leq (1+\gamma)a^2 + (1+\gamma^{-1})b^2, \gamma > 0$ with $\gamma = \frac{\tau_k}{2}$, we get
\begin{align*}
    \norm{\ba_{t+1}}^2 &\leq \tau_k \br{\br{1+\gamma} \norm{\ba_{t}}^2 + \br{1+\gamma^{-1}}\eta_t^2 \norm{\sg_t}^2} \\
    &\leq \frac{(2-\tau_k)}{2} \norm{\ba_{t-1}}^2 + \frac{2}{\tau_k}\eta_t^2 \norm{\sg_t}^2 \\
\end{align*}
Taking expectation on the randomness of the stochastic gradient oracle, and using $\E{\norm{\sg_t}^2} \leq G^2$, we have, 
\begin{align*}
    \E{\norm{\ba_{t+1}}^2} &\leq \frac{(2-\tau_k)}{2} \E{\norm{\ba_{t}}^2} + \frac{2}{\tau_k}\eta_t^2 G^2 \\
\end{align*}
Note that for a fixed $t\leq T$ this recurrence holds with probability at least $1-\delta/T$. Using a union bound, this holds for all $t \in [T]$ with probability at least $1-\delta$. Conditioning on this and using Fact \ref{lem:accumulation2} completes the proof.
\end{proof}
\section{Auxiliary results}
\label{sec:proof2}
We state the result of \citet{stich2018sparsified} in full here.
\thmstich

We now state theorem which uses on the norm bound on stochastic gradients. It follows by directly plugging the fact the \hmx is a $k/d$-contraction in the result of \cite{stich2018sparsified}.
\begin{theorem}
\label{thm:sketchedsgd}
Let $f:\R^d \rightarrow \R$ be a $L$-smooth $\mu$-strongly convex function . Given $T>0$ and $0<k\leq d,0<\delta<1$, Algorithm~\ref{alg:sketchedtopkofsumsgd} on one machine, with access to stochastic gradients such that $\E{\norm{\g}^2} \leq G^2$, with sketch size $\bigO{k \log(dT/\delta)}$ and step size $\eta_t =\frac{1}{t+\xi}$, with $\xi > 1 + \frac{d(1+\beta)}{k(1+\rho)}$, with $\beta>4$ and $\rho = \frac{4 \beta}{(\beta-4)(\beta+1)^2}$, after $T$ steps outputs $\hat \w_T$ such that with probability at least $1-\delta$:
\begin{align*}
    \E{f(\hat \w_T)} - f(\w^*)  &\leq \bigO{\frac{G^2}{\mu T} + \frac{d^2G^2L}{k^2\mu^2T^2} + \frac{d^3G^3}{k^3\mu T^3}}.
\end{align*}
\end{theorem}

\begin{theorem}[(non-convex, smooth)]
\label{thm:ncs}
Let $\{\w_t\}_{t\geq 0}$ denote the iterates of Algorithm~\ref{alg:sketchedtopkofsumsgd} one one machine, on an $L$-smooth function $f: \mathbb{R}^d\to\mathbb{R}$. Assume the stochastic gradients $\g$ satisfy $\mathbb{E}[\g]=\nabla f(\w)$ and $\mathbb{E}[\|\g\|_2^2] \leq G^2$, and use a sketch of size $\mathcal{O}(k\log(dT/\delta))$, for $0\leq \delta \leq 1$. Then, setting  $\eta = 1/\sqrt{T+1}$ with probability at least $1-\delta$: 
\[\min_{t\in [T]}\|\nabla f(\w_t)\|^2 \leq \frac{2f_0}{\sqrt{(T+1)}} + \frac{L G^2}{2\sqrt{T+1}} + \frac{4L^2G^2(1-k/d)}{(k/d)^2(T+1)},\]
where $f_0=f(\w_0) - f^\star$.
\end{theorem}

\begin{theorem}[(convex, non-smooth)]
\label{thm:cns}
Let $\{\w_t\}_{t\geq 0}$ denote the iterates of Algorithm~\ref{alg:sketchedtopkofsumsgd} one one machine, on a convex function $f: \mathbb{R}^d\to\mathbb{R}$. Define $\bar{\w_t}=\frac{1}{T}\sum_{t=0}^{T}\w_t$. Assume the stochastic gradients $\g$ satisfy $\mathbb{E}[\g]=\nabla f(\w)$ and $\mathbb{E}[\|\g\|_2^2] \leq G^2$, and use a sketch of size $\mathcal{O}(k\log(dT/\delta))$, for $0\leq \delta \leq 1$. Then, setting $\eta = 1/\sqrt{T+1}$, with probability at least $1-\delta$: 
\[\mathbb{E}[f(\bar{\w_t})-f^\star] \leq \frac{\|\w_0-\w^\star\|^2}{\sqrt{(T+1)}} + \left(1+\frac{2\sqrt{1-k/d}}{k/d}\right) \frac{G^2}{\sqrt{T+1}}.\]
\end{theorem}

Our high probability bounds of Theorem \ref{thm:sketchedsgd2} can be converted to bounds in expectation, stated below.

\begin{theorem}
\label{thm:sketchedsgdexp}
Let $f:\R^d \rightarrow \R$ be a $L$-smooth $\mu$-strongly convex function. Given $T>0$ and $0<k\leq d,0<\delta<1$, and a $\tau_k$-contraction, Algorithm~\ref{alg:sketchedtopkofsumsgd} one one machine, with sketch size $\bigO{k \log(dT/\delta)}$ and step size $\eta_t =\frac{1}{t+\xi}$, with $\xi > 1 + \frac{1+\beta}{\tau_k(1+\rho)}$, with $\beta>4$ and $\rho = \frac{4 \beta}{(\beta-4)(\beta+1)^2}$ and $\delta = \bigO{\frac{k}{\text{poly}(d)}}$ after $T$ steps outputs $\hat \w_T$ such that
\small
\begin{align*}
    \bbE_{\cA}\E{f(\hat \w_T)} - f(\w^*)  &\leq \bigO{\frac{\sigma^2}{\mu T} + \frac{G^2L}{\tau_k^2\mu^2T^2} + \frac{G^3}{\tau_k^3\mu T^3}}
\end{align*}
\normalsize
\end{theorem}
\begin{proof}
Lemma \ref{lem:contraction} gives that with probability at least $1-\delta$, \hmx is a $k/d$ contraction. We leverage the fact that the elements of countsketch matrix are bounded to convert it to bound in expectation. As in the proof of lemma \ref{lem:contraction}, given $\sg \in \bbR$, the \hmx~algorithm extracts all ${(1/k, \ell^2_2)\text{-heavy}}$ elements from a Count Sketch $S$ of $\sg$. 
Let $\hat \sg$ be the values of all elements recovered from its sketch. For a fixed $k$, we create two sets $H$ (heavy), and $NH$ (not-heavy). All coordinates of $\hat \sg$ with values at least $\frac{1}{k}  \hat \ell_2^2$ are put in $H$, and all others in $NH$, where $\hat \ell_2$ is the estimate of $\|\sg\|_2$ from the Count Sketch. For a $\tau_k$ contraction with probability at least $1-\delta$, we get the following expectation bound.
\begin{align*}
    \bbE_{\cA}\E{\norm{\g - \bar \g}^2} &\leq (1-\delta) \br{1-\tau_k}\norm{\g}^2+  \delta \bigO{\text{poly}(d)}\norm{\g}^2 \\
    & \leq (1-\frac{\tau_k}{2})\norm{\g}^2 
\end{align*}
where the last time follows because we choose $\delta = \frac{\tau_k}{2 \bigO{\text{poly}(d)}}$. Since \hmx is a $k/d$ contraction, we get the expectation bound of $k/2d$ with $\delta = \frac{k}{2d \bigO{\text{poly}(d)}}$

\end{proof}

\section{Sketching}
\label{appendix:sketching}

Sketching gained its fame in the streaming model \citep{muthukrishnan2005data}. A seminal paper by \citet{alon1999space} formalizes the model and delivers a series of important results, among which is the $\ell_2$-norm sketch (later referred to as the AMS sketch).
Given a stream of updates $(a_i, w_i) $ to the $d$ dimensional vector $\g$ (i.e. the $i$-th update is $\g_{a_i} \text{+= } w_i$), the AMS sketch initializes a vector of random signs: 
$s = (s_j)_{j=1}^d, s_j = \pm 1$. On each update $(a_i, w_i)$, it maintains the running sum $S \text{ += } s_{a_i}w_i$, and at the end it reports $S^2$. Note that, if $s_j$ are at least $2$-wise independent, then $E(S^2) = E(\sum_i \g_is_i)^2 = \sum_i \g_i^2 = \|\g\|_2^2$. Similarly, the authors show that $4$-wise independence is enough to bound the variance by $4\|\g\|_2^2$. 
Averaging over independent repetitions running in parallel provides control over the variance, while the median filter (i.e. the majority vote) controls the probability of failure. 
Formally, the result can be summarized as follows: 
AMS sketch, with a large constant probability, finds $\hat \ell_2 = \|\g\|_2 \pm \varepsilon \|\g\|_2$ using only $\bigO{\frac{1}{\varepsilon^2}}$ space. Note that one does not need to explicitly store the entire vector $s$, as its values can be generated on thy fly using $4$-wise independent hashing. 

\begin{definition}\label{def:hh}
Let $\g \in {\mathbb R}^d$. The $i$-th coordinate $\g_i$ of $\g$ is an ${(\alpha_1, \ell_2)\text{-heavy}}$ hitter if $|\g_i| \ge  \alpha_1 \|\g\|_2$. $\g_i$ is an ${(\alpha_2, \ell^2_2)\text{-heavy}}$ hitter if $\g_i^2 \ge  \alpha_2 \|\g \|^2_2$.
\end{definition}

\begin{figure}
\centering
\subfloat[Low level intuition behind the update step of the Count Sketch.]{\includegraphics[width=0.45\textwidth, keepaspectratio]{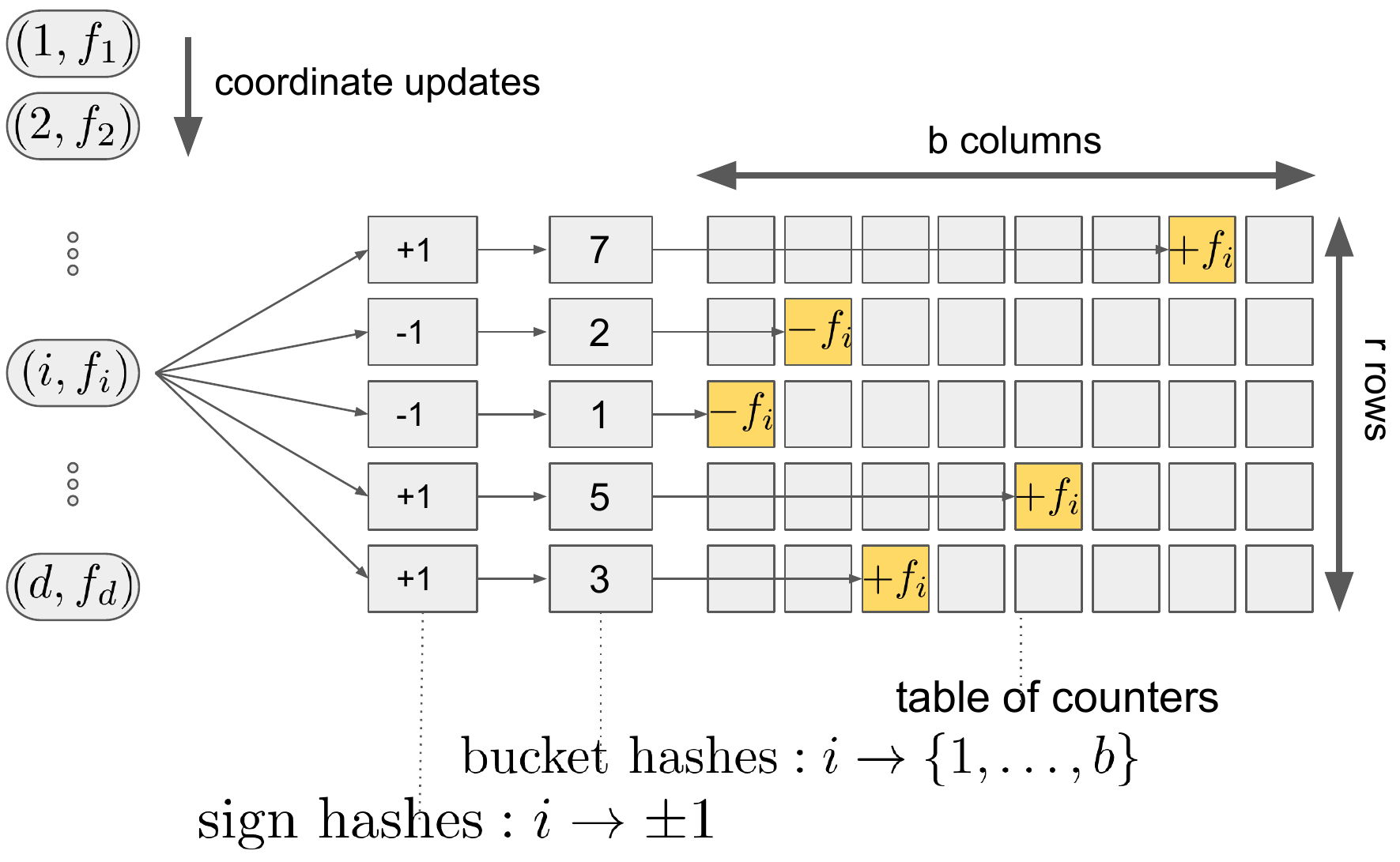}\label{fig:cs_scheme}} 
\hspace{\fill}
\subfloat[Property of mergeability lets the parameter server approximate the heavy coordinates of the aggregate vector]{\includegraphics[width=0.45\textwidth, keepaspectratio]{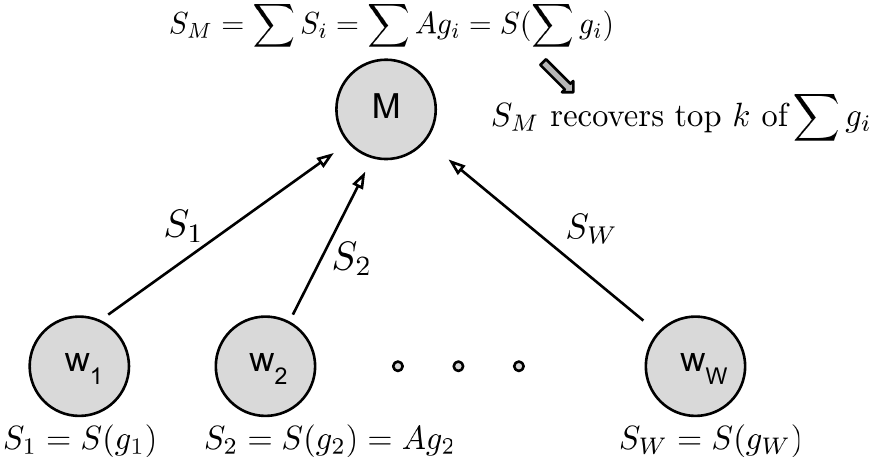}\label{fig:mergeability}} 
\end{figure}

The AMS sketch was later extended by \citet{charikar2002finding} to detect heavy coordinates of the vector (see Definition~\ref{def:hh}).
The resulting Count Sketch algorithm hashes the coordinates into $b$ buckets, and sketches the $\ell_2$ norm of each bucket. 
Assuming the histogram of the vector values is skewed, only a small number of buckets will have relatively large $\ell_2$ norm. Intuitively, those buckets contain the heavy coordinates and therefore all coordinates hashed to other buckets can be discarded. Repeat the same routine independently and in parallel $\bigO{\log_b d}$ times, and all items except the heavy ones will be excluded. Details on how to combine proposed hashing and $\ell_2$ sketching efficiently are presented in Figure~\ref{fig:cs_scheme} and Algorithm~\ref{code:cs}.

Count Sketch finds all $(\alpha, \ell_2)$-heavy coordinates and approximates their values with error $\pm \varepsilon \|\g\|_2$. It does so with a memory footprint of $\bigO{\frac{1}{\varepsilon^2\alpha^2}\log d}$.
We are more interested in finding ${(\alpha, \ell_2^2)\text{-heavy}}$ hitters, which, by an adjustment to Theorem~\ref{thm:cs}, the Count Sketch can approximately find with a space complexity of $\bigO{\frac{1}{\alpha} \log{d}}$, or $\bigO{k \log{d}}$ if we choose $\alpha = \bigO{\frac{1}{k}}$.

Both the Count Sketch and the Count-Min Sketch, which is a similar algorithm presented by \citet{cormode2005improved} that achieves a $\pm \varepsilon \ell_1$ guarantee,
gained popularity in distributed systems primarily due to the mergeability property formally defined by \citet{agarwal2013mergeable}: given a sketch $S(f)$ computed on the input vector $f$ and a sketch $S(g)$ computed on input $g$, 
there exists a function $F$, s.t. $F(S(f),S(g))$ has the same approximation guarantees and 
the same memory footprint as $S(f + g)$. Note that sketching the entire vector can be rewritten as a linear operation $S(f) = Af$, and therefore $S(f+g) = S(f)+S(g)$. We take advantage of this crucial property in \ssgd, since, on the parameter server, the sum of the workers' sketches is identical to the sketch that would have been produced with only a single worker operating on the entire batch.

Besides having sublinear memory footprint and mergeability, the Count Sketch is simple to implement and straight-forward to parallellize, facilitating GPU acceleration \citep{ivkin2018scalable}. 

\citet{charikar2002finding} define the following approximation scheme for finding the list $T$ of the top-$k$ coordinates:
${\forall i \in [d]:  i \in T  \Rightarrow \g_i \ge (1 - \varepsilon) \theta}$  
and ${ \g_i \ge (1 + \varepsilon) \theta \Rightarrow i \in T} $, where $\theta$ is chosen to be the $k$-th largest value of $f$.
\begin{theorem}[\citealp{charikar2002finding}]
\label{thm:cs}
Count Sketch algorithm finds approximate top-$k$ coordinates with probability at least $1-\delta$, in space $O\left(\log{\frac{d}{\delta}} \left(k + \frac{\|\g^{tail}\|_2^2}{(\varepsilon \theta)^2} \right)\right)$, where $\|\g^{tail}\|_2^2 = \sum_{i\notin \text{top k}} \g^2_i$ and $\theta$ is the $k$-th largest coordinate.
\end{theorem}
Note that, if $\theta = \alpha \|\g \|_2$, Count Sketch finds all $(\alpha, \ell_2)$-heavy coordinates and approximates their values with error $\pm \varepsilon \|\g \|_2$. It does so with a memory footprint of $\bigO{\frac{1}{\varepsilon^2\alpha^2}\log d}$.

\begin{algorithm}
    \caption{Count Sketch \citep{charikar2002finding}}
    \label{code:cs}
    \begin{algorithmic}[1]
        \small
        \STATE \textbf{function} init($r$, $c$):
        \STATE ~~~~ init sign hashes $\left\{ s_j\right\}_{j=1}^r$ 
                    and bucket hashes $\left\{ h_j\right\}_{j=1}^r$
        \STATE ~~~~ init $r\times c$ table of counters $S$
    
        \STATE \textbf{function} update($i, f_i$):
        \STATE ~~~~ \textbf{for} $j$ in $1\ldots r$:
        \STATE ~~~~ ~~~~ $S[j, h_j(i)]$ += $s_j(i)f_i$
    
        \STATE \textbf{function} estimate($i$):
        \STATE ~~~~ init length $r$ array estimates
        \STATE ~~~~ \textbf{for} $j$ in $1,\ldots, r$:
        \STATE ~~~~ ~~~~ estimates$[r] = s_j(i) S[j, h_j(i)]$
        \STATE ~~~~ \textbf{return} median(estimates)
        
    \end{algorithmic}
\end{algorithm}

\section{Model Training Details}
\label{app:training}
We train three models on two datasets. For the first two models, we use code from the OpenNMT project \cite{opennmt}, modified only to add functionality for \ssgd. The command to reproduce the baseline transformer results is
\begin{verbatim}
python  train.py -data $DATA_DIR -save_model baseline -world_size 1
        -gpu_ranks 0 -layers 6 -rnn_size 512 -word_vec_size 512
        -batch_type tokens -batch_size 1024 -train_steps 60000
        -max_generator_batches 0 -normalization tokens -dropout 0.1
        -accum_count 4 -max_grad_norm 0 -optim sgd -encoder_type transformer 
        -decoder_type transformer -position_encoding -param_init 0
        -warmup_steps 16000  -learning_rate 1000 -param_init_glorot
        -momentum 0.9 -decay_method noam -label_smoothing 0.1
        -report_every 100 -valid_steps 100
\end{verbatim}
The command to reproduce the baseline LSTM results is
\begin{verbatim}
python  train.py -data $DATA_DIR -save_model sketched -world_size 1
        -gpu_ranks 0 -layers 6 -rnn_size 512 -word_vec_size 512
        -batch_type tokens -batch_size 1024 -train_steps 60000 
        -max_generator_batches 0 -normalization tokens -dropout 0.1
        -accum_count 4 -max_grad_norm 0 -optim sgd -encoder_type rnn 
        -decoder_type rnn -rnn_type LSTM -position_encoding -param_init 0 
        -warmup_steps 16000 -learning_rate 8000 -param_init_glorot
        -momentum 0.9 -decay_method noam -label_smoothing 0.1
        -report_every 100 -valid_steps 100
\end{verbatim}
We run both models on the WMT 2014 English to German translation task, preprocessed with a standard tokenizer and then shuffled.

The last model is a residual network trained on CIFAR-10. We use the model from the winning entry of the DAWNBench competition in the category of fastest training time on CIFAR-10 \cite{dawnbench}. We train this model with a batch size of 512, a learning rate varying linearly at each iteration from 0 (beginning of training) to 0.4 (epoch 5) back to 0 (end of training). We augment the training data by padding images with a 4-pixel black border, then cropping randomly back to 32x32, making 8x8 random black cutouts, and randomly flipping images horizontally. We use a cross-entropy loss with L2 regularization of magnitude $0.0005$.

Each run is carried out on a single GPU -- either a Titan X, Titan Xp, Titan V, Tesla P100, or Tesla V100.

\section{Additional experiments}

\subsection{MNIST}

\begin{figure}[t]
\centering
   \subfloat[{\footnotesize log-log plot of training and test error against number of iterations of the average iterate for SVM trained on one class as positive and the rest as negative (1-v-all). For simplicity, we only show the plot for one class.}]{\includegraphics[width=0.50\textwidth]{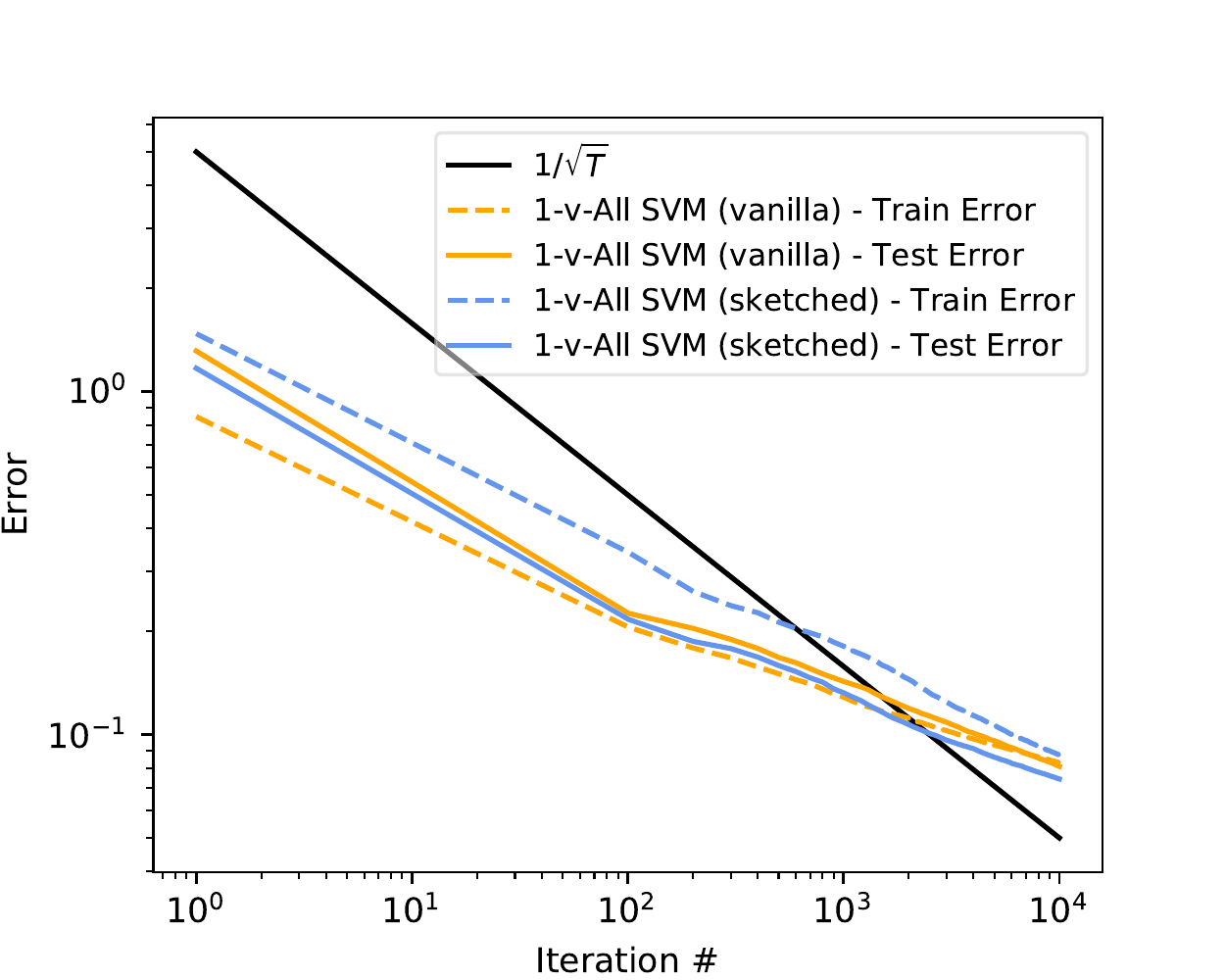}
      \label{fig:mnistsvm}}
      \hspace{\fill}
   \subfloat[log-log plot of training and test error of the  number of iterations for regularized logistic regression. The regularization parameter was fixed as $0.01$.]{\includegraphics[width=0.45\textwidth]{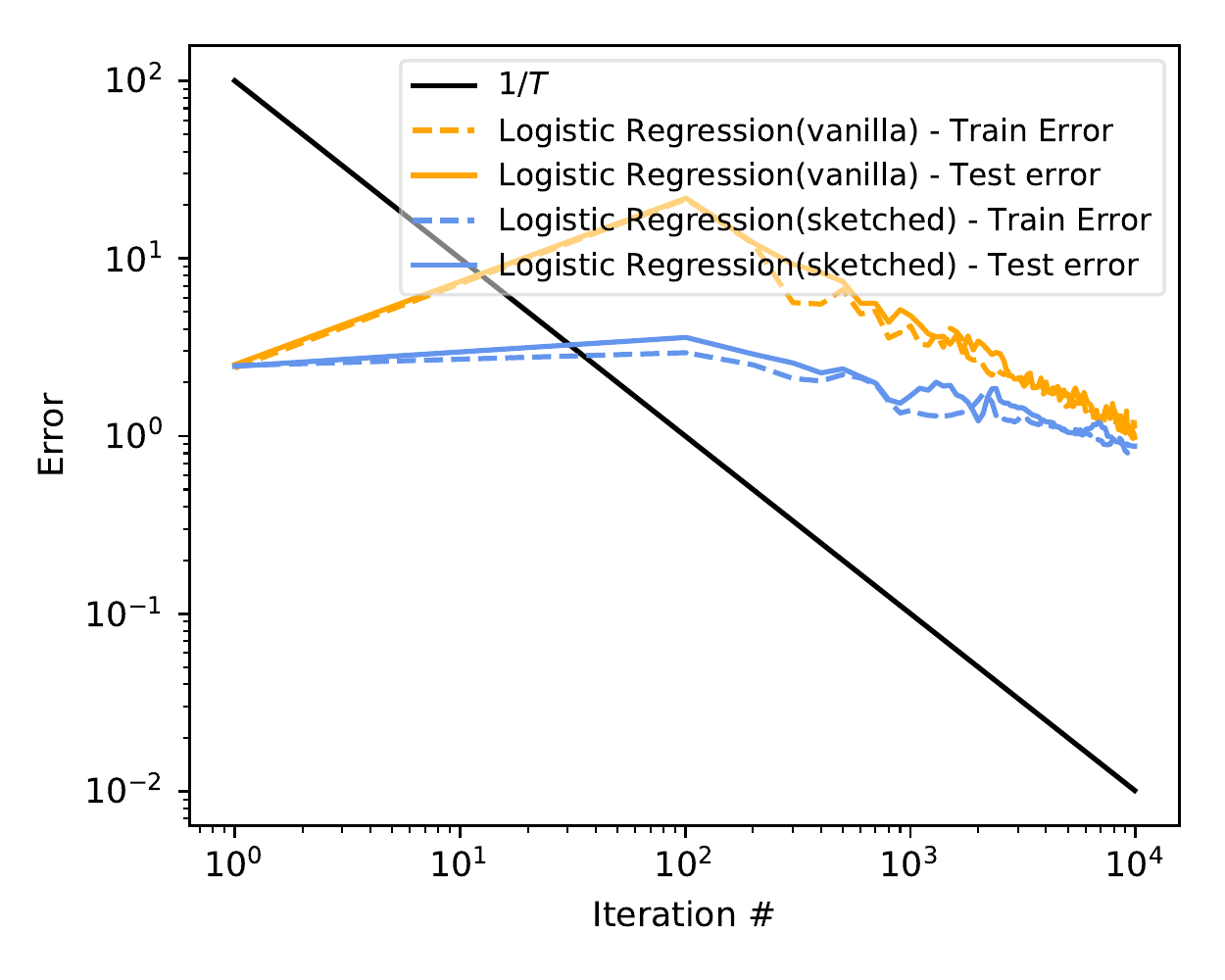}
         \label{fig:mnistlogreg}}
\end{figure}
We train vanilla and sketched counterparts of two simple learning models: Support vector machines(SVM) and $\ell_2$ regularized logistic regression on MNIST dataset. These are examples of optimizing non-smooth convex function and strongly convex smooth function respectively.
We also compare against the theoretical rates obtained in Theorems \ref{thm:cns} and \ref{thm:stronglyconvexsmooth}. The sketch size used in these experiments is size 280 (40 columns and 7 rows), and the parameters $k$ and $P$ are set as, $k=10,P=10$, giving a compression of around 4; the number of workers is 4. Figure \ref{fig:mnistsvm} and \ref{fig:mnistlogreg} shows the plots of training and test errors of these two models. In both the plots, we see that the train and test errors decreases with $T$ in the same rates for vanilla and sketched models. However, these are conservative compared to the theoretical rate suggested.

\end{document}